\documentclass[runningheads]{llncs}

\usepackage{eccv}

\usepackage{eccvabbrv}

\usepackage{graphicx}
\usepackage{booktabs}
\usepackage{amsmath}
\usepackage{amssymb}
\usepackage{multirow}
\usepackage{xcolor}
\usepackage{tikz}
\usetikzlibrary{positioning, arrows.meta, decorations.pathreplacing, calc}
\usepackage[section]{placeins}

\usepackage[accsupp]{axessibility}  % Improves PDF readability for those with disabilities.

\usepackage{hyperref}

\usepackage{orcidlink}

\begin{document}

% ---------------------------------------------------------------
\title{VisCritic: Visual State Comparison as Process Reward for GUI Agents} 

\titlerunning{VisCritic}

\author{Jiachen Qian}

\authorrunning{Qian}

\institute{City University of Hong Kong\\
\email{72510756@cityu-dg.edu.cn}}

\maketitle

\begin{abstract}
  GUI agents powered by vision-language models show strong potential for automating digital tasks, yet frequently fail in long-horizon scenarios due to the absence of step-level verification. Existing process reward models verify actions through textual reasoning alone, missing the visual nature of GUI state changes. We introduce \textbf{VisCritic}, a visual process reward framework that verifies agent actions by directly comparing pre-action and post-action screenshots in visual feature space. VisCritic employs a Siamese vision transformer to extract change-aware representations, coupled with an Action-Aware Critic Head that jointly evaluates action success, task progress, and error type. A critic-training data construction pipeline generates weakly supervised samples from existing trajectories without additional human labels for critic training. Experiments and offline analyses across five benchmarks demonstrate that VisCritic serves as a plug-and-play enhancement for diverse GUI agents, generally improving benchmark metrics while providing visual diagnostic cues.
\end{abstract}

\section{Introduction}
\label{sec:intro}

Graphical User Interface (GUI) agents~\cite{seeclick,showui,uground} powered by multimodal large language models~\cite{gpt4,llava,qwen2vl,internvl2} automate digital tasks by interpreting screenshots and generating executable actions such as clicking, typing, and scrolling. Despite rapid advances in visual grounding~\cite{seeclick,guig1,segui} and action prediction, these agents remain fragile in long-horizon tasks, where a single erroneous action can cascade into task failure~\cite{osworld,backtrackagent}. Recent studies further show that screenshot-based and multimodal agents can be compromised by visual perturbations or poisoned visual memories~\cite{pennywise,visualinception}, reinforcing the need for verification mechanisms grounded in visual state evidence.

\begin{figure}[t]
  \centering
  \includegraphics[width=0.86\linewidth]{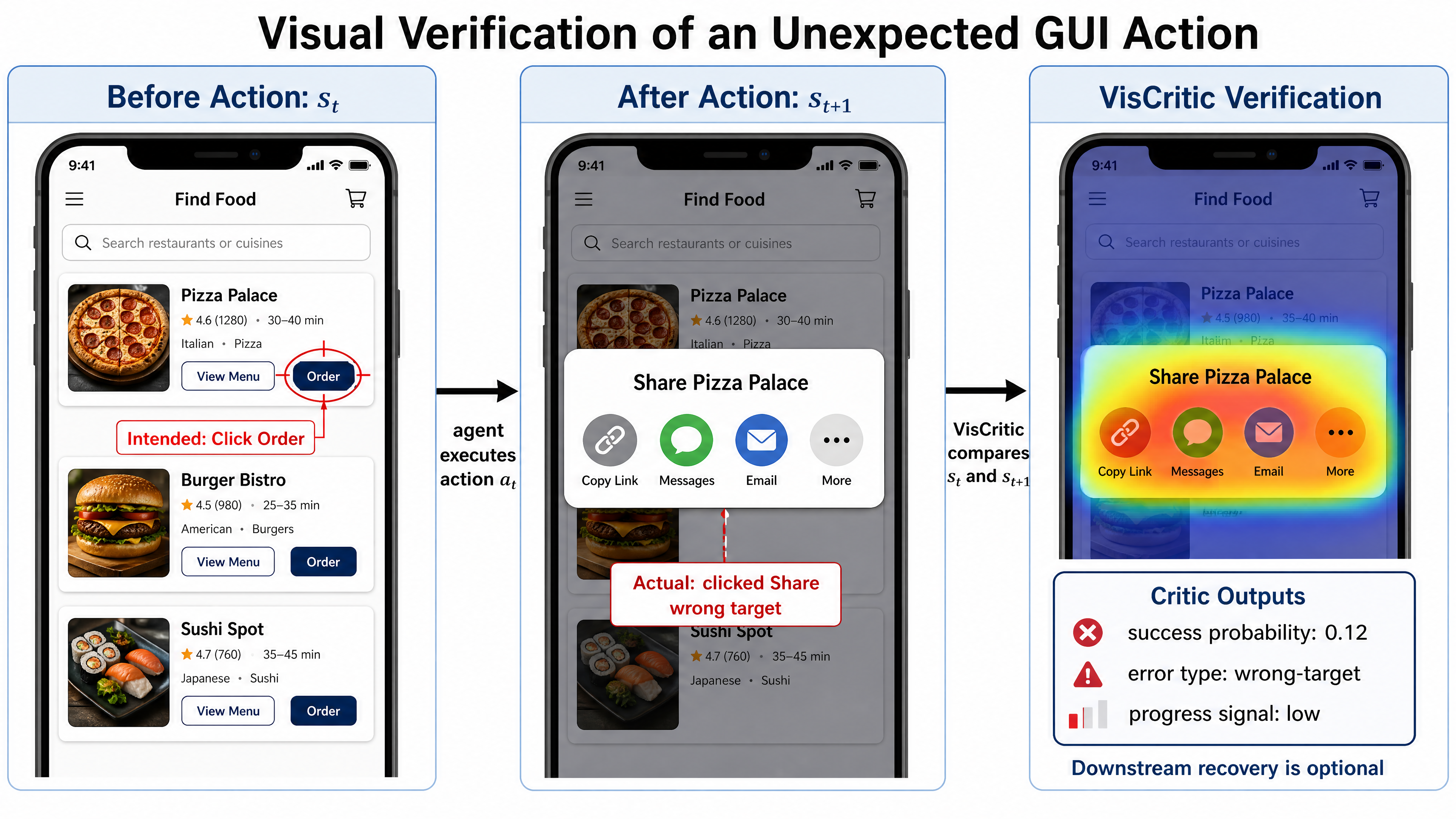}
  \caption{\textbf{Motivation example} (schematic illustration). A GUI agent accidentally clicks ``Share'' instead of ``Order'' (left$\to$middle). VisCritic detects the error by comparing screenshots: the attention map highlights the unexpected popup, and the critic predicts low success with ``wrong target'' classification (right).}
  \label{fig:motivation}
\end{figure}

A fundamental limitation of current GUI agents is the absence of reliable \emph{step-level verification}: after executing an action, the agent has no mechanism to confirm whether the action produced its intended visual effect. Consider the example in \cref{fig:motivation}: the agent intends to click an ``Order'' button but instead clicks an adjacent element. Without verification, the agent proceeds based on a false assumption, inevitably leading to failure.

Recent work has begun to address this gap through process reward models~\cite{guipra,guidnav} and error detection mechanisms~\cite{backtrackagent,guicriticr1}. However, these approaches share a critical limitation: \emph{their verification signal is largely mediated through textual reasoning, tool calls, or structured checklists}. This verification paradigm is fundamentally misaligned with the visual nature of GUI interactions, where action outcomes manifest as pixel-level state changes---a button changes color, a dialog appears, a page scrolls to reveal new content.

\begin{figure}[t]
  \centering
  \includegraphics[width=0.92\linewidth]{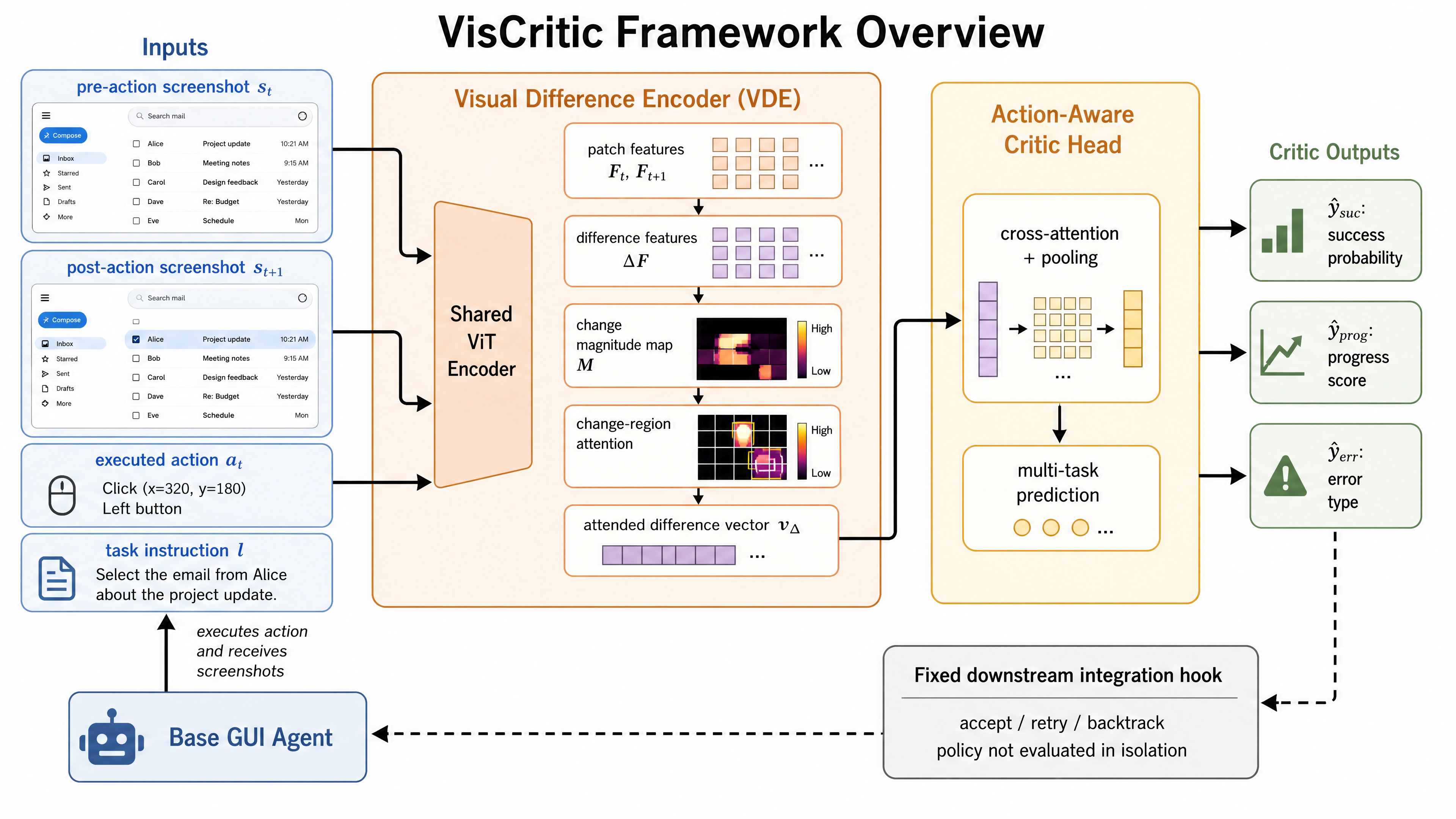}
  \caption{\textbf{Overview of the VisCritic framework.} Given a pre-action screenshot $s_t$ and post-action screenshot $s_{t+1}$, the Visual Difference Encoder (VDE) extracts patch-level semantic differences via a shared ViT encoder, computes a change magnitude map, and applies change region attention to produce the attended difference vector $\mathbf{v}_\Delta$. The Action-Aware Critic Head fuses $\mathbf{v}_\Delta$ with the action $a_t$ and task instruction $l$ to predict action success, task progress, and error type. The critic output can be passed to a fixed downstream integration hook for accept, retry, or backtrack decisions.}
  \label{fig:framework}
\end{figure}

In this paper, we propose \textbf{VisCritic}, a visual process reward framework that bridges this gap by directly comparing pre-action and post-action screenshots to verify GUI agent actions (\cref{fig:framework}). Our key observation is that \emph{visual differences between the screen before and after an action are often an informative signal when action outcomes manifest visually}. VisCritic realizes this idea with a Siamese Visual Difference Encoder, an Action-Aware Critic Head, and a critic-training data construction pipeline that weakly supervises training from existing trajectories. It operates as a plug-and-play inference-time module for screenshot-based GUI agents that expose action traces and pre-/post-action screenshots, without modifying the agent architecture or requiring additional training of the agent itself.

Evaluation across five benchmarks spanning web, mobile, and desktop environments demonstrates improvements in most tested interactive settings and outperforms text-based PRM baselines in the majority of evaluated settings. VisCritic also provides qualitative diagnostic cues through change-region attention maps. Our contributions are summarized as follows:
\begin{enumerate}
    \item We identify the fundamental mismatch between textual verification and visual GUI state changes, and propose visual state comparison as a more natural and effective process reward signal.
    \item We design VisCritic, a plug-and-play visual critic framework with a Siamese-based Visual Difference Encoder and multi-task Critic Head that jointly predicts action success, task progress, and error type.
    \item We introduce a weakly supervised data pipeline that builds critic-training samples from existing trajectories without additional human labels.
    \item Extensive experiments and offline analyses on five benchmarks show that VisCritic generally improves diverse GUI agents and outperforms text-based verification approaches in the majority of evaluated settings.
\end{enumerate}

\section{Related Work}
\label{sec:related}

\textbf{GUI Agents.}
GUI agents automate digital tasks by perceiving screen content and executing interface actions. Early approaches rely on structured representations such as HTML or accessibility trees~\cite{mind2web}, but recent work has shifted toward purely visual agents that operate directly on screenshots. SeeClick~\cite{seeclick} demonstrates that GUI grounding is the performance bottleneck. UGround~\cite{uground} proposes scalable synthetic data for visual grounding. ShowUI~\cite{showui} unifies vision-language-action modeling with UI-guided token selection. Concurrently, LLM-based agents~\cite{webagent,appagent,mobileagent} demonstrate autonomous GUI navigation. Despite these advances, many current GUI agents still rely primarily on single-pass action prediction and lack reliable step-level verification of executed action outcomes.

\textbf{Process Reward Models for GUI Agents.}
PRMs provide intermediate supervision to guide agent behavior during multi-step tasks~\cite{prm_verify}. GUI-PRA~\cite{guipra} introduces a process reward agent with dynamic memory and adaptive UI perception, explicitly addressing the ``lost in the middle'' problem and lack of UI changing awareness; however, its verification operates through textual tool calls. CRAFT-GUI~\cite{craftgui} designs task-specific reward functions by integrating rule-based verification with model-predicted evaluation.

More recently, GUI-Shepherd~\cite{guishepherd} provides dense step-by-step process supervision trained on 52K human-annotated interactions, serving both as an RL reward provider and inference-time verifier. Web-Shepherd~\cite{webshepherd} proposes the first PRM specifically for web agents, using structured subgoal checklists. ADMIRE~\cite{admire} addresses the reward fidelity--density trade-off by anchoring trajectories to dynamically distilled milestones. While these PRMs achieve notable improvements, their verification signal is largely mediated through textual state descriptions, tool calls, or structured checklists---thus inheriting the modality gap between textual verification and the visual nature of GUI state changes. VisCritic provides a complementary visual verification signal that could be combined with textual PRMs for multi-modal process supervision.

\textbf{Error Detection and Recovery.}
BacktrackAgent~\cite{backtrackagent} introduces a four-component framework for error detection and recovery through backtracking, where the verifier is rule-based and the judger is a trained VLM. GUI-Critic-R1~\cite{guicriticr1} takes a ``look before you leap'' approach, pre-judging actions before execution. The CVPR 2025 work on VLM self-correction~\cite{vlm_selfcorrect} explores whether VLMs can correct their own grounding errors through feedback. Minitap~\cite{minitap} achieves 100\% success on AndroidWorld via a multi-agent architecture with task decomposition and deterministic post-validation, but its per-action-type validators are limited to action types with programmatically verifiable outcomes. VisCritic generalizes verification to a broad range of GUI action types by operating in visual feature space, complementing deterministic validators.

\textbf{Visual Change Detection, World Models, and Tree Search.}
Visual change detection has been extensively studied in remote sensing~\cite{changedetection_survey,fcef} and video surveillance, where Siamese architectures~\cite{siamfc} have proven effective for identifying meaningful differences between temporally separated images. We apply this paradigm to a novel context: using visual state differences as a reward signal for agent decision-making. Concurrent GUI world models---MobileDreamer~\cite{mobiledreamer}, CUWM~\cite{cuwm}, and Dyna-Mind~\cite{dynamind}---address state \emph{prediction} rather than verification. VisCritic is complementary: while world models predict \emph{what will happen}, VisCritic verifies \emph{what actually happened} without depending on the fidelity of learned dynamics.

Tree search methods such as LATS~\cite{lats} and Agent Alpha~\cite{agentalpha} improve \emph{action selection} by exploring multiple trajectories via MCTS. VisCritic addresses the orthogonal problem of \emph{action verification}---confirming whether a chosen action produced its intended visual effect.

\section{Method}
\label{sec:method}

\subsection{Problem Formulation}

We formulate GUI task automation as a sequential decision process. At each time step $t$, the agent observes a screenshot $s_t$, receives a task instruction $l$, and generates an action $a_t$ (\eg, \texttt{CLICK(x,y)}, \texttt{TYPE("text")}, \texttt{SCROLL(direction)}). The environment executes $a_t$ and transitions to a new state, producing screenshot $s_{t+1}$. The task is complete when the agent reaches a goal state or exceeds a maximum number of steps.

\textbf{Visual Critic} is defined as a function $\mathcal{C}: (s_t, a_t, s_{t+1}, l) \rightarrow (\hat{y}_{\text{suc}}, \hat{y}_{\text{prog}}, \hat{y}_{\text{err}})$ that takes the before-after screenshot pair $(s_t, s_{t+1})$, the executed action $a_t$, and the task instruction $l$, and outputs:
\begin{itemize}
    \item $\hat{y}_{\text{suc}} \in [0, 1]$: action success probability,
    \item $\hat{y}_{\text{prog}} \in [-1, 1]$: task progress score,
    \item $\hat{y}_{\text{err}} \in \{$success, no-op, wrong-target, page-error, timeout$\}$: error type.
\end{itemize}
When $\hat{y}_{\text{suc}} < \gamma$ (a success threshold), the critic can optionally expose its verification result to downstream recovery policies, such as retrying, backtracking, or requesting a textual recovery suggestion.

\subsection{Visual Difference Encoder}
\label{sec:vde}

The Visual Difference Encoder (VDE) is the core component responsible for extracting meaningful visual changes between consecutive screenshots. Rather than relying on pixel-level image differencing, which is sensitive to rendering noise, animation artifacts, and minor layout shifts, VDE operates in a learned semantic feature space.

\subsubsection{Siamese Feature Extraction.}
Given screenshots $s_t$ and $s_{t+1}$, a shared-weight ViT~\cite{vit} encoder $\mathcal{E}_\theta$ extracts patch-level features:
\begin{equation}
    F_t = \mathcal{E}_\theta(s_t) \in \mathbb{R}^{P \times d}, \quad F_{t+1} = \mathcal{E}_\theta(s_{t+1}) \in \mathbb{R}^{P \times d},
\end{equation}
where $P$ is the number of visual patches and $d$ is the feature dimension. The Siamese architecture~\cite{siamfc} ensures that both screenshots are encoded in the same feature space, enabling meaningful comparison.

\subsubsection{Difference Feature Computation.}
We compute the patch-level difference features:
\begin{equation}
    \Delta F = F_{t+1} - F_t \in \mathbb{R}^{P \times d}.
\end{equation}
Additionally, we compute a \emph{change magnitude map} $M \in \mathbb{R}^{P}$:
\begin{equation}
    M_i = \| \Delta F_i \|_2, \quad i = 1, \ldots, P,
\end{equation}
which indicates the degree of semantic change at each patch position.

\subsubsection{Change Region Attention.}
We introduce change region attention to focus on regions with meaningful UI changes:
\begin{equation}
    \alpha_i = \frac{\exp(M_i / \beta)}{\sum_{j=1}^{P} \exp(M_j / \beta)},
\end{equation}
where $\beta$ is a learnable temperature parameter. The attended difference representation is:
\begin{equation}
    \mathbf{v}_{\Delta} = \sum_{i=1}^{P} \alpha_i \cdot \Delta F_i \in \mathbb{R}^{d}.
\end{equation}

The full VDE output consists of three components: (1) the global attended difference vector $\mathbf{v}_{\Delta}$, (2) the patch-level difference features $\Delta F$ for fine-grained analysis, and (3) the change magnitude map $M$ for interpretability.

\subsection{Action-Aware Critic Head}
\label{sec:critic_head}

The Critic Head integrates visual difference features with action and task context to produce multi-task predictions. We encode the action $a_t$ and task instruction $l$ using the text encoder of the backbone MLLM:
\begin{equation}
    \mathbf{h}_a = \text{TextEnc}(a_t), \quad \mathbf{h}_l = \text{TextEnc}(l).
\end{equation}

The fused representation is obtained via cross-attention, where the patch-level difference features serve as queries and the textual context provides keys and values, combined with spatial self-attention over $\Delta F$:
\begin{align}
    \mathbf{z}_{\text{cross}} &=
    \text{Pool}\!\bigl(\text{CrossAttn}(Q{=}\Delta F,\; K{=}[\mathbf{h}_a; \mathbf{h}_l],\; V{=}[\mathbf{h}_a; \mathbf{h}_l])\bigr), \notag\\
    \mathbf{z} &= \mathbf{z}_{\text{cross}} +
    \text{Pool}\!\bigl(\text{SpatialAttn}(\Delta F)\bigr),
\end{align}
where both branches are projected to the same hidden dimension and $\text{Pool}(\cdot)$ denotes change-magnitude-weighted pooling using $\alpha$ from \cref{sec:vde}.
The multi-task outputs are:
\begin{align}
    \hat{y}_{\text{suc}} &= \sigma(\text{MLP}_{\text{suc}}(\mathbf{z})), \\
    \hat{y}_{\text{prog}} &= \tanh(\text{MLP}_{\text{prog}}(\mathbf{z})), \\
    \hat{y}_{\text{err}} &= \text{softmax}(\text{MLP}_{\text{err}}(\mathbf{z})).
\end{align}

When $\hat{y}_{\text{suc}} < \gamma$, downstream agents may optionally use $\mathbf{z}$ and the predicted error type to drive retry, backtracking, or recovery hooks. In this work we focus on the core accept/reject verification; recovery policy design is outside the main quantitative scope.

\subsection{Critic-Guided Agent Loop}
\label{sec:agent_loop}

VisCritic integrates with existing GUI agents through two complementary mechanisms:

\subsubsection{Best-of-N Selection (Pre-Execution, Optional).}
When a visual state predictor is available (\eg, MobileDreamer~\cite{mobiledreamer}), the agent generates $N$ candidate actions and VisCritic scores each based on the predicted post-action state $\hat{s}_{t+1}^{(a)}$, rendered as simplified layout images. VisCritic selects the highest-scoring action (details and a text-only fallback analysis are in the supplementary material):
\begin{equation}
    a_t^* = \arg\max_{a \in \{a_t^{(1)}, \ldots, a_t^{(N)}\}} \mathcal{C}(s_t, a, \hat{s}_{t+1}^{(a)}, l)_{\text{suc}}.
\end{equation}

\subsubsection{Post-Execution Verification.}
After executing the selected action, VisCritic verifies the actual outcome by comparing $s_t$ and the real $s_{t+1}$. This is the primary verification mode and does not require any state prediction---it operates on the \emph{actual} visual outcome. If $\hat{y}_{\text{suc}} < \gamma$, the agent can enter an optional recovery policy that retries, backtracks, or requests a recovery suggestion. In our interactive experiments, low-confidence steps are handled by the same fixed retry/recovery hook across compared settings; thus task-SR gains should be interpreted as the effect of VisCritic's verification signal under this fixed integration protocol, rather than as an isolated evaluation of recovery-policy design. Post-execution verification is the recommended default mode, as it uses actual post-execution screenshots and avoids learned state prediction.

\subsection{Training Pipeline}
\label{sec:training}

\subsubsection{Annotation-Free Data Construction.}

\begin{figure}[t]
  \centering
  \includegraphics[width=0.96\linewidth]{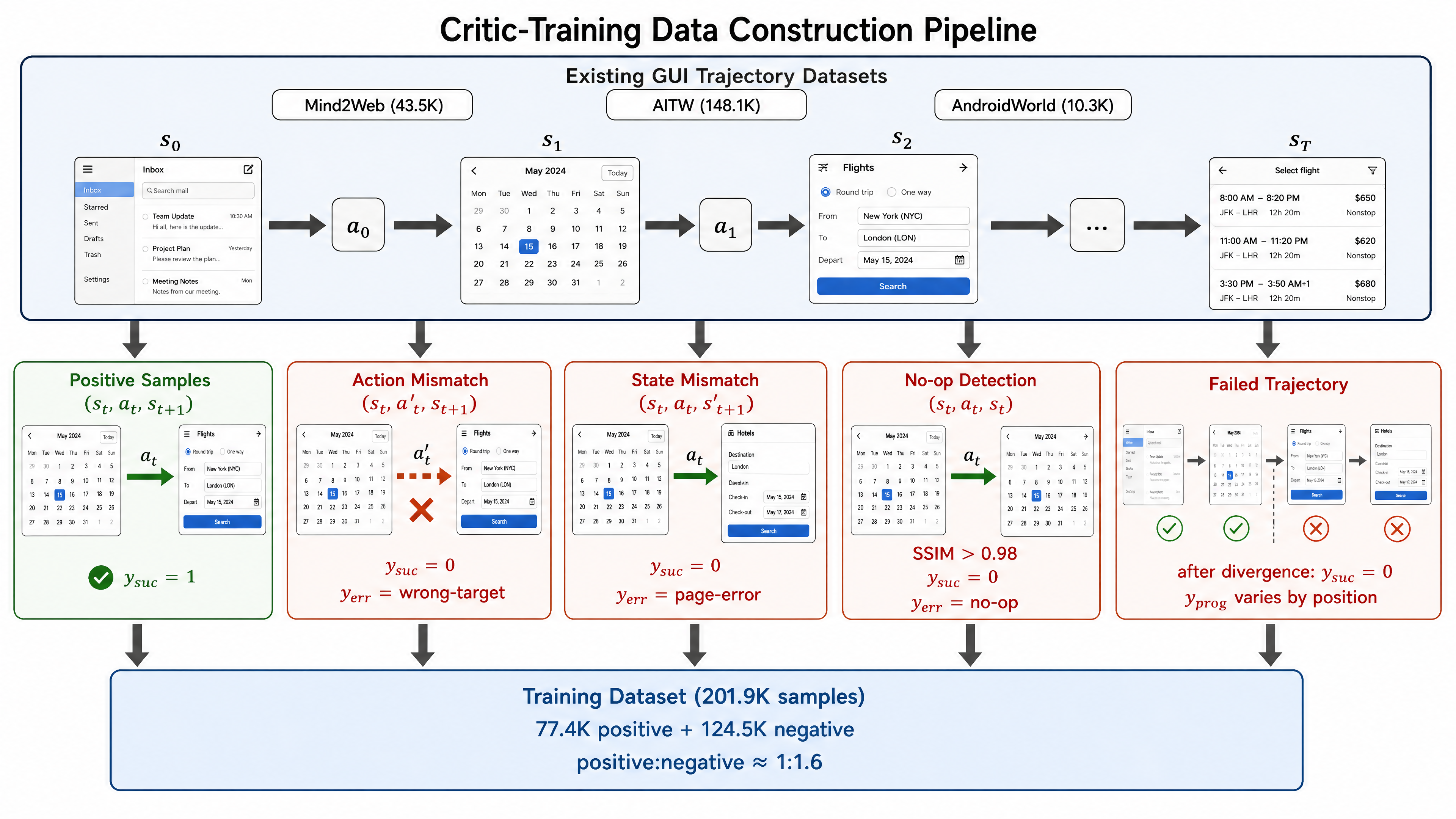}
  \caption{\textbf{Critic-training data construction pipeline.} From existing GUI trajectory datasets, we automatically generate five types of training samples: positive samples from successful trajectories, and four types of negatives (action mismatch, state mismatch, no-op, failed trajectory) without additional human labels for critic training.}
  \label{fig:data_pipeline}
\end{figure}

A key advantage of VisCritic is that critic-training data can be automatically constructed from existing GUI trajectory datasets without additional human labels for critic training. As illustrated in \cref{fig:data_pipeline}, we define five types of training samples:

\textbf{Positive samples} are drawn from successful trajectories, where consecutive tuples $(s_t, a_t, s_{t+1})$ are assigned a weak positive proxy label $y_{\text{suc}} = 1$.

\textbf{Negative samples} are generated through four perturbation strategies:
\begin{enumerate}
    \item \textbf{Action mismatch}: Replace $a_t$ with a random action $a'$ from the same trajectory while keeping $s_t$ and $s_{t+1}$ fixed, creating cases where the visual change does not correspond to the described action.
    \item \textbf{State mismatch}: Replace $s_{t+1}$ with a screenshot from a different trajectory, creating cases where the visual outcome is unexpected.
    \item \textbf{No-op detection}: Pair $s_t$ with $s_{t+1} \approx s_t$ (SSIM $>$ 0.98) to train detection of ineffective actions.
    \item \textbf{Failed trajectory sampling}: Draw tuples from failed trajectories. Steps after the estimated divergence point are assigned a weak negative proxy label $y_{\text{suc}} = 0$; the divergence point is estimated by comparing against the closest successful trajectory.
\end{enumerate}

\textbf{Error-type labels} ($y_{\text{err}}$) are derived deterministically from the perturbation strategy (81.4\% agreement with 500 manually annotated cases). \textbf{Progress scores} are trajectory-derived heuristic labels: $y_{\text{prog}} = (t{+}1)/T$ for successful steps, and $y_{\text{prog}} = -k/K$ after the divergence point in failed trajectories. Details are in the supplementary material.

\subsubsection{Multi-Phase Training.}

\textbf{Phase 1: Contrastive Pre-training.} The VDE is pre-trained with an InfoNCE loss~\cite{infonce,simclr}:
\begin{equation}
    \mathcal{L}_{\text{contrast}} = -\log \frac{\exp(\text{sim}(\mathbf{v}_\Delta, \mathbf{h}_a^+) / \tau_c)}{\sum_{j} \exp(\text{sim}(\mathbf{v}_\Delta, \mathbf{h}_a^{(j)}) / \tau_c)},
\end{equation}
where $\mathbf{h}_a^+$ is the correct action feature. For samples with a reliable action-region mask, an auxiliary localization loss (KL divergence) encourages attention $\alpha$ to concentrate on UI elements affected by the action, using a binary mask $\mathbf{m}$ from action coordinates:
\begin{equation}
    \mathcal{L}_{\text{loc}} = \sum_{i=1}^{P} \hat{\alpha}_i \log \frac{\hat{\alpha}_i}{\alpha_i},
\end{equation}
with $\hat{\alpha}_i = m_i / \sum_j m_j$ when $\sum_j m_j>0$. Samples without a reliable action-region mask skip $\mathcal{L}_{\text{loc}}$. The Phase~1 objective is $\mathcal{L}_{\text{P1}} = \mathcal{L}_{\text{contrast}} + 0.3 \cdot \mathcal{L}_{\text{loc}}$ for masked samples. The localization mask is used \emph{only during training}; at inference, attention is computed solely from the learned change magnitude map $M$.

\textbf{Phase 2: Multi-Task Fine-Tuning.} The full model is trained end-to-end with a multi-task loss~\cite{multitask_uncertainty}:
\begin{equation}
    \mathcal{L} = \lambda_1 \mathcal{L}_{\text{BCE}}(\hat{y}_{\text{suc}}, y_{\text{suc}}) + \lambda_2 \mathcal{L}_{\text{MSE}}(\hat{y}_{\text{prog}}, y_{\text{prog}}) + \lambda_3 \mathcal{L}_{\text{CE}}(\hat{y}_{\text{err}}, y_{\text{err}}),
\end{equation}
where $\lambda_1, \lambda_2, \lambda_3$ are loss weights.

\section{Experiments}
\label{sec:experiments}

\subsection{Experimental Setup}

\subsubsection{Benchmarks.}
We evaluate on five benchmarks: \textbf{Mind2Web}~\cite{mind2web} (cross-task test split, 252 web tasks, step SR), \textbf{AndroidWorld}~\cite{androidworld} (116 total tasks; we evaluate on a 34-task application-disjoint held-out split, live emulator), \textbf{OSWorld}~\cite{osworld} (369 desktop tasks), \textbf{WebArena}~\cite{webarena} (812 web tasks), and \textbf{AITW}~\cite{aitw} (715K Android episodes, action accuracy). For AITW, which is an offline benchmark, we report a logged-screenshot consistency diagnostic using recorded pre-/post-action screenshots; these numbers should not be interpreted as live inference-time gains for arbitrary candidate actions.

\subsubsection{Base Agents.}
To demonstrate plug-and-play compatibility, we apply VisCritic to multiple base agents: SeeClick~\cite{seeclick}, ShowUI~\cite{showui}, CogAgent~\cite{cogagent}, and an agent based on Qwen2.5-VL~\cite{qwen2vl}.

\subsubsection{Baselines.}
We compare against: (1)~base agents without any critic, (2)~vanilla Best-of-N with self-consistency, (3)~GUI-PRA~\cite{guipra} (text-based PRM), (4)~GuidNav~\cite{guidnav} (text PRM with search), (5)~BacktrackAgent~\cite{backtrackagent} (text-based error detection), and (6)~GUI-Critic-R1~\cite{guicriticr1} (pre-execution critic). All baselines are re-implemented using publicly released code and evaluated under our unified protocol where applicable, with AITW treated as an offline diagnostic.\footnote{Re-implemented numbers may differ from original publications due to our unified evaluation protocol (screenshot-only input, unified action space). Details in the supplementary material.} We note that GUI-Shepherd~\cite{guishepherd} has not released its training data or model weights. Web-Shepherd~\cite{webshepherd} releases a web-focused PRM and WebPRM Collection, but its checklist-based reward formulation, annotation protocol, and web-agent setting differ from our screenshot-pair visual critic protocol. A parameter-, data-, and protocol-controlled comparison is therefore non-trivial, so we restrict baselines to methods whose released artifacts permit faithful re-implementation under our unified protocol. For GUI-PRA, we use a Qwen2.5-VL-7B supervisor to match the parameter budget of other baselines. The Pixel-Critic baseline (\cref{tab:visual_vs_textual}) uses the same Critic Head architecture but replaces the VDE with raw pixel-level difference maps ($|s_{t+1} - s_t|$ in RGB space), ensuring a fair comparison that isolates the contribution of learned semantic features.

\textbf{Baseline fairness.} VisCritic (InternViT-6B + Qwen2.5-7B, $\sim$13B total) uses a larger backbone than some baselines. To control for model capacity, the Text-Critic in \cref{tab:visual_vs_textual} uses the \emph{same} Qwen2.5-7B backbone and identical Critic Head, differing only in modality; VisCritic's advantage (84.6\% vs.\ 66.2\% F1 in \cref{tab:visual_vs_textual}) supports improvement from the visual comparison paradigm rather than model scale. In the controlled AndroidWorld critic-quality setting, a smaller SigLIP-400M~\cite{siglip} backbone ($\sim$7.4B total) remains stronger than the text-based critic baselines reported in the same setting (80.3\% F1 and 23.1\% task SR; see supplementary material).

\subsubsection{Implementation Details.}
VisCritic uses InternViT-6B~\cite{internvl2,vit} with LoRA~\cite{lora} adapters (rank=16) and Qwen2.5-7B text backbone. Training data ($\sim$202K samples) is constructed from Mind2Web (43.5K), AITW (148.1K), and AndroidWorld (10.3K) with $\sim$1:1.6 positive-to-negative ratio. OSWorld and WebArena are \emph{not} in the training data; results on these benchmarks represent zero-shot cross-benchmark generalization (and additionally cross-platform for OSWorld's desktop environment). All evaluation uses held-out test splits with strict de-duplication (details in supplementary material). Phase~1 trains the VDE for 5 epochs (lr=1e-4, $\lambda_{\text{loc}}{=}0.3$); Phase~2 fine-tunes end-to-end for 10 epochs (lr=2e-5, $\lambda_1{=}1.0$, $\lambda_2{=}0.5$, $\lambda_3{=}0.5$). We set $\gamma{=}0.5$; unless otherwise stated, main results use post-execution verification, while $N{=}3$ is used only in optional Best-of-$N$ analyses. All experiments use 8$\times$A100 GPUs.

\subsection{Main Results}

\begin{table}[tb]
  \caption{Performance (\%) across benchmarks (task SR for AndroidWorld/\allowbreak OSWorld/\allowbreak WebArena, step SR for Mind2Web, action accuracy for AITW). AITW is an offline logged-screenshot consistency diagnostic rather than a live interaction setting. ``Interactive Avg.''\ is the arithmetic mean over Mind2Web, AndroidWorld, OSWorld, and WebArena only. Best in \textbf{bold}. $^\dagger$Not all baselines shown; some require agent-specific hooks unavailable in released codebases (see supplementary).}
  \label{tab:main_results}
  \centering
  \resizebox{\textwidth}{!}{
  \begin{tabular}{l|ccccc|c}
    \toprule
    Method & Mind2Web & AndroidWorld & OSWorld & WebArena & AITW & Interactive Avg. \\
    \midrule
    SeeClick & 33.4 & 14.7 & 6.8 & 12.4 & 62.1 & 16.8 \\
    \quad + GUI-PRA & 36.2 & 17.1 & 8.2 & 14.6 & 64.5 & 19.0 \\
    \quad + BacktrackAgent & 36.9 & 17.9 & 9.0 & 15.3 & 64.2 & 19.8 \\
    \quad + GUI-Critic-R1 & 37.2 & 18.3 & 9.3 & 15.7 & \textbf{64.8} & 20.1 \\
    \quad + \textbf{VisCritic (Ours)} & \textbf{37.6} & \textbf{19.1} & \textbf{9.5} & \textbf{16.4} & 64.5 & \textbf{20.7} \\
    \midrule
    ShowUI & 43.5 & 19.8 & 10.3 & 17.6 & 66.7 & 22.8 \\
    \quad + GUI-PRA & 46.2 & 22.1 & 12.0 & 19.7 & 68.8 & 25.0 \\
    \quad + \textbf{VisCritic (Ours)} & \textbf{48.9} & \textbf{24.8} & \textbf{12.8} & \textbf{23.2} & \textbf{69.8} & \textbf{27.4} \\
    \midrule
    CogAgent$^\dagger$ & 41.2 & 17.5 & 8.9 & 15.1 & 64.8 & 20.7 \\
    \quad + \textbf{VisCritic (Ours)} & \textbf{44.7} & \textbf{20.8} & \textbf{10.6} & \textbf{18.4} & \textbf{67.1} & \textbf{23.6} \\
    \midrule
    Qwen2.5-VL Agent & 47.3 & 24.6 & 14.2 & 22.8 & 70.4 & 27.2 \\
    \quad + GUI-PRA & 50.1 & 27.0 & 16.3 & 24.9 & 72.5 & 29.6 \\
    \quad + GuidNav & 51.2 & 27.8 & 17.0 & 26.1 & 73.0 & 30.5 \\
    \quad + GUI-Critic-R1 & 51.8 & 28.2 & \textbf{17.4} & 26.5 & 73.3 & 31.0 \\
    \quad + \textbf{VisCritic (Ours)} & \textbf{54.1} & \textbf{29.8} & 17.1 & \textbf{29.4} & \textbf{74.2} & \textbf{32.6} \\
    \bottomrule
  \end{tabular}
  }
\end{table}

\cref{tab:main_results} presents the main results. We highlight three key observations: (1)~Excluding the offline AITW diagnostic, VisCritic improves every base agent on the large majority of interactive evaluated settings, demonstrating its generality as an inference-time plug-in under our tested protocol; in one interactive case (Qwen2.5-VL on OSWorld), a text-based critic marginally outperforms, reflecting scenarios where textual verification may have a natural advantage in text-heavy desktop environments. (2)~The improvement is particularly pronounced on interactive web benchmarks (WebArena, Mind2Web), supporting that step-level visual verification is most valuable when error accumulation is severe. (3)~Among benchmark--agent combinations where released text-based baselines could be faithfully re-implemented under our unified protocol, VisCritic outperforms text-based critic baselines in the majority of settings, supporting that visual state comparison generally provides a reliable verification signal complementary to textual reasoning about GUI changes. Notably, the improvement generally correlates with base agent capability---Qwen2.5-VL gains +5.4 pts in interactive average and ShowUI gains +4.6 pts---though the relationship is not strictly monotonic (CogAgent gains +2.9 pts, possibly due to its dual-encoder architecture limiting plug-and-play integration).

\textbf{Statistical significance.} All results are averaged over 3 independent runs; standard deviations range from 0.3 pts (AITW, largest offline diagnostic set) to 1.4 pts (AndroidWorld, smallest held-out split). On Mind2Web, WebArena, and AITW, improvements over the best text-based baseline are statistically significant ($p < 0.01$, paired bootstrap). On OSWorld, $p < 0.05$. On AndroidWorld, the improvement is directionally consistent but does not reach conventional significance ($p \approx 0.08$) due to the small held-out split (34 tasks); a larger evaluation pool would provide stronger statistical power, and the per-task win rate (VisCritic wins on 20/34 tasks vs.\ best baseline) is directionally consistent. We use paired bootstrap with task/episode-level resampling where applicable; p-values are nominal and uncorrected for multiple comparisons.

\subsection{Critic Quality Analysis}

\cref{tab:critic_quality} evaluates the intrinsic quality of action verification.

\begin{table}[tb]
  \caption{Action verification quality on AndroidWorld (Qwen2.5-VL base). We report precision, recall, and F1 (\%) for action success prediction.}
  \label{tab:critic_quality}
  \centering
  \begin{tabular}{l|ccc}
    \toprule
    Method & Precision & Recall & F1 \\
    \midrule
    GUI-PRA (Text) & 73.2 & 67.8 & 70.4 \\
    BacktrackAgent (Text) & 75.6 & 64.3 & 69.5 \\
    \textbf{VisCritic (Visual)} & \textbf{87.1} & \textbf{83.4} & \textbf{85.2} \\
    \bottomrule
  \end{tabular}
\end{table}

\subsection{Visual vs. Textual Critic Comparison}

To isolate the benefit of visual state comparison, we conduct a controlled experiment where all critics share identical Critic Head architecture but differ in how they assess UI state change: (1)~\textbf{Text-Critic} uses textual descriptions of expected changes; (2)~\textbf{Pixel-Critic} uses raw pixel difference maps; (3)~\textbf{VisCritic} uses learned VDE semantic features.

\begin{table}[tb]
  \caption{Visual vs.\ textual critic comparison: action verification F1 (\%) by action category on held-out AndroidWorld + Mind2Web.}
  \label{tab:visual_vs_textual}
  \centering
  \begin{tabular}{l|ccc|c}
    \toprule
    Critic & Visual-Only & Semantic & No-Op & Overall \\
    \midrule
    Text-Critic & 58.3 & 74.1 & 62.7 & 66.2 \\
    Pixel-Critic & 71.6 & 63.8 & 78.4 & 70.5 \\
    \textbf{VisCritic} & \textbf{84.3} & \textbf{85.6} & \textbf{84.2} & \textbf{84.6} \\
    \bottomrule
  \end{tabular}
\end{table}

\cref{tab:visual_vs_textual} presents the results. Three findings stand out: (1)~Text-Critic performs poorly on \emph{visual-only} changes (\eg, button highlight, icon state toggle, color transitions) that are difficult to describe textually, achieving only 58.3\% F1 compared to VisCritic's 84.3\%. (2)~Pixel-Critic struggles with \emph{semantic} changes (\eg, page navigation, content loading) where pixel-level differences are large but uninformative, scoring 63.8\% F1 versus VisCritic's 85.6\%. (3)~VisCritic achieves balanced high performance across all categories, supporting that learned semantic difference features capture both fine-grained visual changes and high-level state transitions effectively. Notably, VisCritic's No-Op detection (84.2\%) significantly outperforms Text-Critic (62.7\%), as no-op cases exhibit a distinctive minimal-change pattern in the VDE feature space that is difficult to express textually but easy to detect visually.

\subsection{Ablation Study}

\cref{tab:ablation} isolates the contribution of each VisCritic component.

\begin{table}[tb]
  \caption{Ablation study on VisCritic components. Task success rate (\%) on AndroidWorld with ShowUI base agent.}
  \label{tab:ablation}
  \centering
  \begin{tabular}{l|c}
    \toprule
    Variant & Success Rate \\
    \midrule
    Full VisCritic & 24.8 \\
    \quad w/o Change Region Attention & 23.4 \\
    \quad w/o Multi-task Head (success only) & 24.1 \\
    \quad w/o Contrastive Pre-training & 22.0 \\
    \quad w/o Post-execution Verification & 21.4 \\
    \quad Pixel-level Diff (replace VDE) & 21.2 \\
    \bottomrule
  \end{tabular}
\end{table}

Key observations: (1)~Contrastive pre-training is the most impactful training component ($-$2.8 pts), supporting the importance of learning discriminative visual difference representations before multi-task fine-tuning. (2)~Removing post-execution verification ($-$3.4 pts) weakens the fixed recovery loop, showing that verification against actual screenshots is critical under our integration protocol; pre-execution selection alone is insufficient. (3)~Replacing VDE with pixel-level differences ($-$3.6 pts) validates that learned semantic features are essential; raw pixel differences are too sensitive to rendering noise. (4)~Change region attention contributes +1.4 pts, focusing the model on relevant UI changes. (5)~The multi-task head provides a modest +0.7 pts over success-only prediction; while auxiliary tasks benefit training, the success prediction objective dominates. Additional ablations on backbone choice, Best-of-N candidates, and loss weights are in the supplementary material.

\subsection{Qualitative Analysis}

\begin{figure}[t]
  \centering
  \includegraphics[width=0.92\linewidth]{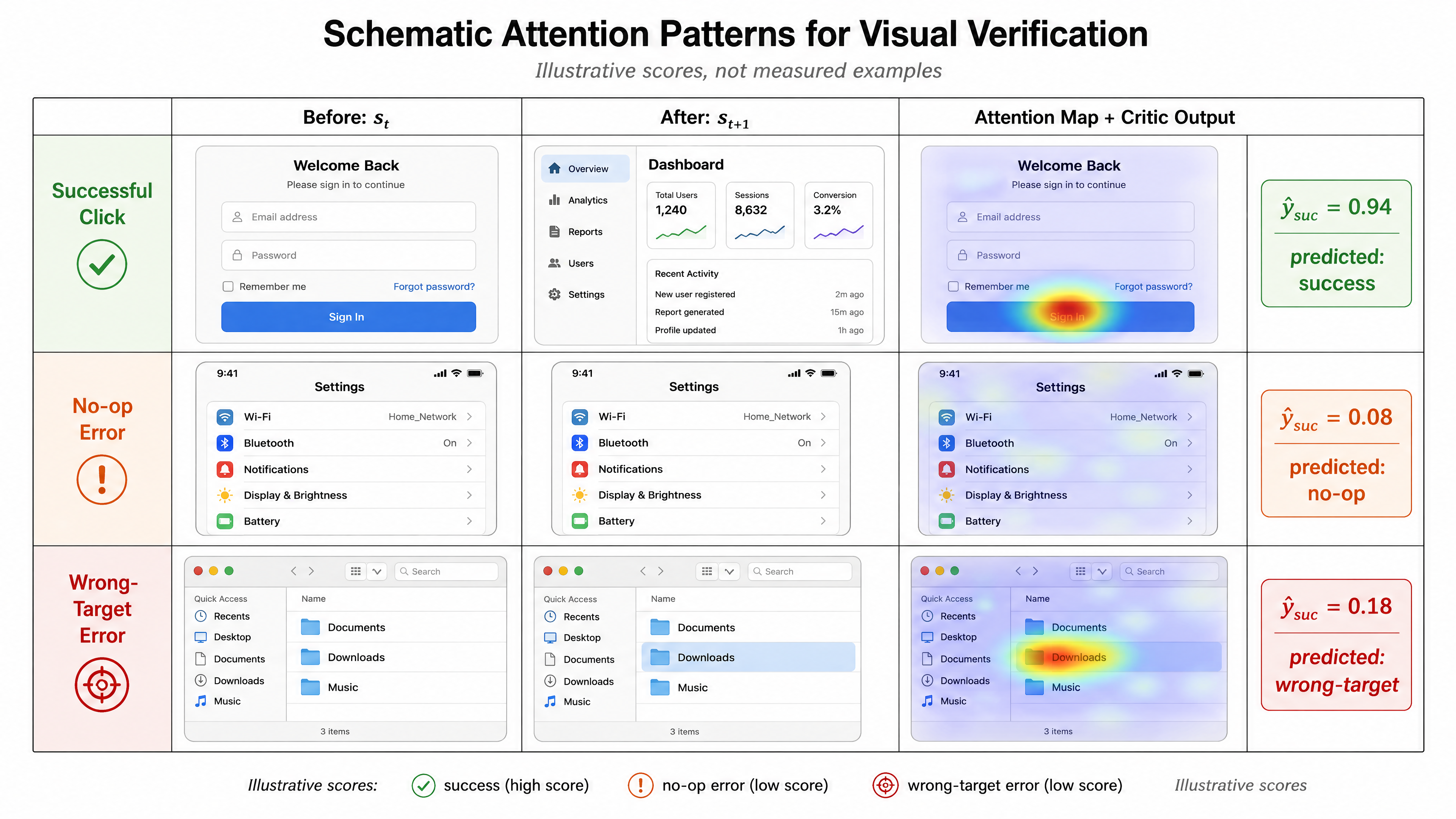}
  \caption{\textbf{Conceptual illustration of change region attention patterns.} (top) For successful clicks, attention is expected to concentrate on the clicked element ($\hat{y}_{\text{suc}} > 0.9$). (middle) For no-op errors, attention disperses uniformly ($\hat{y}_{\text{suc}} < 0.15$). (bottom) For wrong-target errors, attention focuses on an unintended region ($\hat{y}_{\text{suc}} \approx 0.2$). UI mockups are schematic; category-level verification performance is reported in \cref{tab:visual_vs_textual}.}
  \label{fig:qualitative}
\end{figure}

\cref{fig:qualitative} schematically illustrates three characteristic attention patterns consistent with the corresponding verification outcomes. For \textbf{successful clicks}, the change region attention concentrates on the clicked element and its visual response ($\hat{y}_{\text{suc}} > 0.9$). For \textbf{no-op errors}, the attention disperses uniformly due to the absence of meaningful visual change ($\hat{y}_{\text{suc}} < 0.15$). For \textbf{wrong-target errors}, the attention focuses on an unintended changed region ($\hat{y}_{\text{suc}} \approx 0.2$).

\subsection{Cross-Agent Transferability and Overhead}

As shown in \cref{tab:main_results}, a single VisCritic model trained once transfers to four architecturally distinct agents without any agent-specific adaptation, supporting that visual state verification is largely agent-agnostic under our tested integration protocol.

VisCritic adds 6.2B parameters (InternViT-6B with LoRA adapters plus the Critic Head) when the Qwen2.5-7B text backbone is shared with the base agent; for agents using a different LLM backbone (\eg, SeeClick, CogAgent), an additional text encoder is required. While this is a significant overhead, we note that: (1)~the critic runs independently and can be deployed on a separate GPU; (2)~post-execution mode adds only 66ms per step (15.7\% overhead on A100-80GB with FP16, comprising 52ms for the Siamese ViT forward pass and 14ms for the Critic Head), negligible compared to typical 1--3s environment interaction latency; (3)~the observed 1.7--6.8\% absolute metric improvements may justify the additional compute in accuracy-critical settings. For resource-constrained settings, a SigLIP-400M backbone reduces overhead to 35ms while retaining 80.3\% critic F1 (see supplementary material). Additional analyses including progress score evaluation, error-type classification, robustness to dynamic visual changes, and pre- vs.\ post-execution comparison are provided in the supplementary material.

\section{Conclusion}
\label{sec:conclusion}

We presented VisCritic, a visual process reward framework for GUI agents that verifies action outcomes through direct visual state comparison. By introducing a Siamese-based Visual Difference Encoder with change region attention, VisCritic captures meaningful UI changes at the semantic level rather than relying on textual descriptions. Our critic-training data construction pipeline enables scalable weak supervision without additional human labels for critic training. Extensive experiments and offline analyses demonstrate that VisCritic serves as an effective plug-and-play enhancement for diverse GUI agents, outperforming evaluated text-based verification approaches on the large majority of comparable benchmark--agent combinations. The change-region attention maps further offer a qualitative view of which visual changes influence verification.

\textbf{Limitations and future work.} VisCritic inherits the visual understanding limitations of its backbone ViT encoder, particularly for fine-grained text changes or subtle UI state transitions (\eg, a counter incrementing from ``3'' to ``4''). Task-irrelevant visual dynamics (animations, ads, asynchronous loads) can degrade verification accuracy; while the learned semantic features and change region attention provide meaningful robustness, extreme viewport and resolution changes remain challenging. The critic-training data construction pipeline relies on heuristic weak labels, including divergence point estimation via trajectory alignment, which may inject label noise (sensitivity analysis in the supplementary material). Recovery policies, while architecturally compatible with VisCritic outputs, have not been quantitatively evaluated in isolation.

Several promising directions remain for future work: (1)~\textbf{Lightweight variants}: while our default InternViT-6B backbone achieves the best accuracy, the SigLIP-400M variant retains 94\% of the critic F1 at 15$\times$ fewer vision parameters, making deployment practical on consumer GPUs; exploring distillation or architecture-specific optimizations could further reduce the footprint. (2)~\textbf{Hybrid verification}: combining VisCritic with deterministic validators~\cite{minitap} for action types with programmatically verifiable outcomes (\eg, text input) could improve coverage on such action types, as the two approaches are naturally complementary. (3)~\textbf{Integration with tree search}: using VisCritic as a learned value function within MCTS frameworks~\cite{lats,agentalpha} could provide more grounded state assessments than text-based evaluators. (4)~\textbf{Multi-modal process supervision}: combining VisCritic's visual signal with textual PRM scores~\cite{guishepherd,webshepherd} for fused multi-modal verification.

\bibliographystyle{splncs04}
\bibliography{main}
\end{document}